\documentclass[10pt,twocolumn,letterpaper]{article}

\usepackage{cvpr}
\usepackage{times}
\usepackage{epsfig}
\usepackage{graphicx}
\usepackage{amsmath}
\usepackage{amssymb}

\usepackage{algorithm}
\usepackage[noend]{algpseudocode}
\usepackage{booktabs}
\usepackage{multirow}
\usepackage{makecell}
\usepackage{xparse}
\usepackage{bbm}

\usepackage[pagebackref=true,breaklinks=true,letterpaper=true,colorlinks,bookmarks=false]{hyperref}

\cvprfinalcopy 

\def\cvprPaperID{3} 

\ifcvprfinal\fi
\begin{document}

\title{Fast forwarding Egocentric Videos by Listening and Watching}
\author{Vinicius S. Furlan$^1$, Ruzena Bajcsy$^2$, Erickson R. Nascimento$^{1,2}$ \\
 Universidade Federal de Minas Gerais (UFMG), Brazil$^1$\\
 University of California Berkeley, USA$^2$\\
{\tt\small \{viniciusfurlan, erickson\}@dcc.ufmg.br}\\
{\tt\small bajcsy@eecs.berkeley.edu}
}

\maketitle

\begin{abstract}

The remarkable technological advance in well-equipped wearable devices is pushing an increasing production of long first-person videos. However, since most of these videos have long and tedious parts, they are forgotten or never seen. Despite a large number of techniques proposed to fast-forward these videos by highlighting relevant moments, most of them are image based only.  Most of these techniques disregard other relevant sensors present in the current devices such as high-definition microphones. In this work, we propose a new approach to fast-forward videos using psychoacoustic metrics extracted from the soundtrack. These metrics can be used to estimate the annoyance of a segment allowing our method to emphasize moments of sound pleasantness. The efficiency of our method is demonstrated through qualitative results and quantitative results as far as of speed-up and instability are concerned.

\end{abstract}

\section{Introduction}

Thanks to the recent technology advances, a flood of low-cost wearable cameras equipped with high-quality sensors is reaching the consumers. The low-cost of these cameras and the large number of sharing and storing websites are popularizing the use of these well-equipped devices. People are increasingly logging their daily routines, generating massive amounts of egocentric videos rich in visual and sound information. However, since most parts of the egocentric videos are tedious to watch, long egocentric videos are doomed to be forgotten.


Video Summarization and Semantic-aware Hyperlapse are two popular approaches for reducing the size of egocentric videos. Although Video Summarization techniques can find the meaningful moments of a video, they return only disconnected fragments of the whole video~\cite{yao2016highlight, cote2016, lu2013story}. Semantic-aware Hyperlapse works~\cite{lai2017semantic, Ramos2016, Silva2016, Silva2018CVPR, Silva2018}, on the other hand, can identify relevant moments while preserving the timeline of the video. Despite the remarkable advances on video summarization and hyperlapse works, they are still restricted to visual information. Virtually all hyperlapse and summarization methods disregard an important piece of information provided by additional sensors like microphones, the sound.
The sound is informative and may provide important clues of the context of a scene and be used to assign importance to segments of a video based on metrics like loudness and annoyance.



In this work, we propose a novel methodology that combines the psychoacoustic metrics with visual features to fast-forward first-person videos. By also considering the sound, we can measure the relevance of a frame using the psychoacoustic annoyance metrics and avoid selecting unpleasure moments like a segment with a crowd or a noisy street. We present quantitative and qualitative results that show the ability of our method in fast-forwarding a video using the sound and images.

\section{Related Work}


We can roughly divide the methods of selecting  meaningful moments in long-duration videos into two categories: Video Summarization and Semantic-Aware Hyperlapse.

Video Summarization techniques try to identify the most informative segments of a video and create a compact summary of these moments. There are many ways to classify a segment as informative, e.g., user's preference~\cite{yao2016highlight}, abnormal behaviors~\cite{cote2016}, and story-telling~\cite{lu2013story}. For instance, Lee \etal~\cite{lee2015sumspeak} propose a method to summarize a video using both face detection and speech recognition. The authors start by identifying regions in the images containing human faces. Then, they generated coefficients based on the face regions and used these coefficients as features in a classification step. After extracting the audio from the speaker in the video, the face detection and speaker verification results are used to summarize the video.

\begin{figure*}[t!]
\centering
\includegraphics[width=0.87\textwidth]{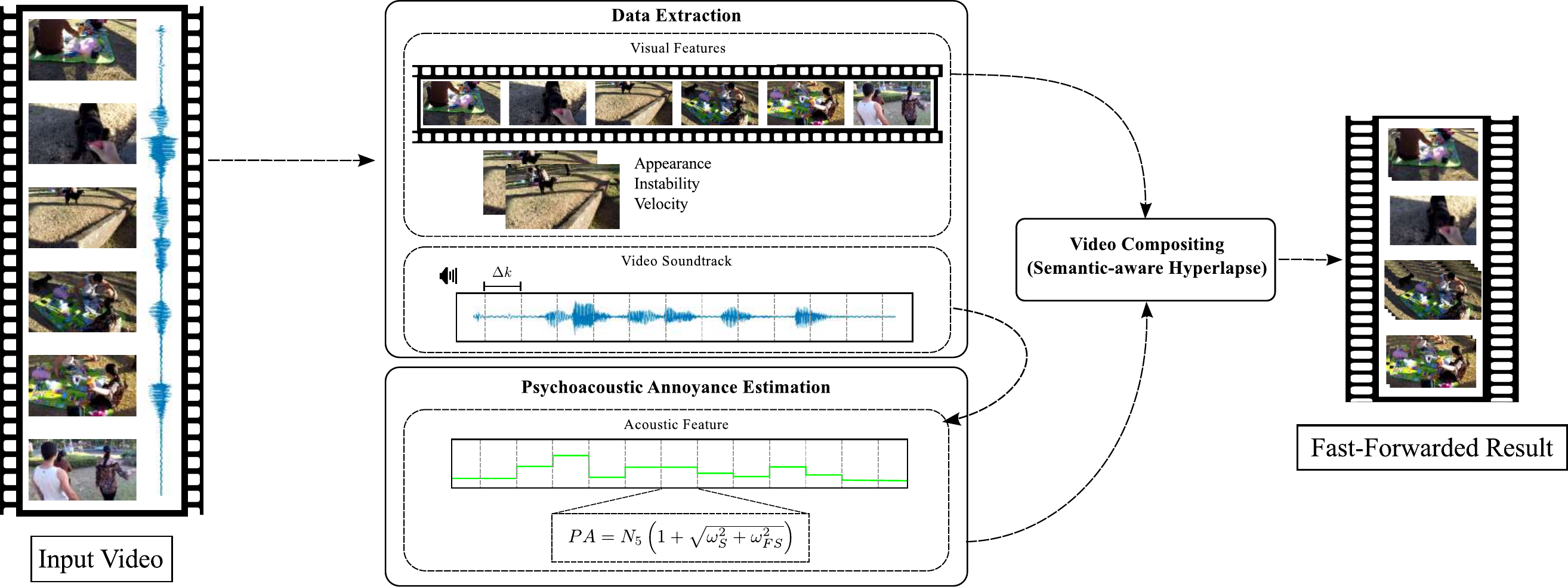}
\caption{Diagram illustrating the three main steps of methodology. After segmenting the soundtrack of the input video into slices of size $\Delta k$,  we compute the PA metric (green curve). The PA value is a semantic score assigned to each segment. The semantic score is used to create a relevant profile of the video. This profile is used in the video compositing step to select the relevant frames.}
\label{fig:methodology}
\end{figure*}

Although video summarization techniques achieved remarkable results creating summaries, they do no generate a pleasant experience for the user, once they miss the temporal continuity. Semantic-aware Hyperlapse techniques, like the works of Silva~\etal~\cite{Silva2016, Silva2018, Silva2018CVPR}, Ramos~\etal~\cite{Ramos2016} and Lai~\etal~\cite{lai2017semantic}, preserve the temporal continuity and the smoothness of a video while emphasizing relevant segments. However, these Semantic-aware Hyperlapse techniques rely only on visual information, they overlook the information provided by other sensors such as the sound.

Several recent studies have been trying to combine sound and sight. A recent and representative approach is the work of Owens~\etal~\cite{owens2018audiovisual}. The authors propose a self-supervised way to learn a multi-sensory representation that jointly models audio and visual information from a video. The learned representation is then used to predict sound source localization, audio-visual action recognition, and on/off-screen audio separation. Arandjelovic~\etal~\cite{arandjelovic2017look} present a system that can learn the semantic information of a scene by looking and listening to unlabelled videos.

Similar to the works of Owens~\etal and Arandjelovic~\etal, in this work we combine sound and visual data. We present an approach that uses psychoacoustic metrics extracted from the sound of a scene in conjunction with visual information to fast-forward first-person videos.
\section{Methodology}
\label{sec:methodology}

Our fast-forward method consists of three primary steps, outlined in Figure~\ref{fig:methodology}: data extraction, the psychoacoustic annoyance estimation and video compositing.

\paragraph{Data Extraction.}
In this step, we extract the soundtrack from a video of length $n$ and segment it into slices of lower length $\Delta k$ resulting in the following set $\mathcal{S} = \{s_1, s_2, s_3, ..., s_m\}$, where $s_{i}$ is the $i$-th slice of the soundtrack and $m=\frac{n}{\Delta k}$.

As stated, our methodology combines acoustical and visual information. Thus, after collecting the sound data, we compute the instability, appearance, and velocity between frames. The instability metric is estimated by the average distance Focus of Expansion (FOE) to the center of the frame. To estimate appearance, we compute the Earth Mover's Distance between the color histogram of the frames. At last, the velocity factor is given by the difference between the average magnitude of the optical flows of the frames along with the optical flows of the whole video. All these metrics and the psychoacoustic annoyance metric (computed from the sound) are joined in a linear combination that is used in the video compositing step.

\paragraph{Psychoacoustic Annoyance estimation.}

In the context of first-person videos, auditory stimulus is a relevant factor for perceiving  the environment. Zwicker's metric~\cite{zwicker2013psychoacoustics} is a popular method for estimating the sound impression for a human listener. It is based on four psychoacoustic measurements:
\begin{itemize}
    \item {\it Fluctuation} and {\it Roughness}: A complex environment has multiple frequencies sounds that constructively and destructively interfere with each other creating modulation. Fluctuation and roughness are two measurements of the modulation of a signal over the time. The fluctuation was designed to work with $20$ up to modulations per second; roughness describes sounds with modulations range from $20$ to $300$ times per second. A modulated signal is considerably more unpleasant when having a higher roughness and fluctuation. In this work, we used the Roughness metric proposed by Daniel and Weber~\cite{daniel1997psychoacoustical};

	\item {\it Loudness} and {\it Sharpness}: While loudness takes into account the distributions of critical bands in the human hearing, sharpness is a function of the spectral composition. The sharpness metric is estimated by a weighted sum of specific loudness levels in different bands. A sound with high sharpness is more annoying. The loudness is a psychological phenomenon. Different from a sound level that is a physical measurement, the loudness was developed based on human subject studies in persons with normal hearing. People in these studies listened to a tone at frequency $f$ Hz and a particular dB level, a second tone was then played at a different frequency. The level of this second tone would be altered until it sounded equally as loud as the $f$ Hz tone. 
\end{itemize}

Zwicker proposes to compute the psychoacoustic annoyance (PA) as a function of sharpness (S), loudness (N), fluctuation (F), and roughness (R) as:
\begin{eqnarray}
&& PA = N_5 \left( 1 + \sqrt{\omega_S^2 + \omega_{FS}^2} \right),\\
&& \omega_S = \mathbbm{1}{[S > 0]}\times(1.75-S) \log(N_5+10), \\
&& \omega_{FS} = 2.78 \times N_5^{-0.4} \times (0.4 F + 0.6 R),
\end{eqnarray}
\noindent where $N_5$ is the 95th percentile of loudness and $\mathbbm{1}{[X]}$ is the indicator function, having the value $1$ if $X$ is true and the value $0$ otherwise.

After computing the median values of Roughness, Fluctuation, Sharpness and Loudness for each slice $s_i$, we estimate the PA value. We used the shape function $f(PA)=\lambda e^{-\lambda PA}$ to compute the semantic score since the video compositing step relates high values of the score with semantic information. In our case, high PA values represent irrelevant information. The parameter $\lambda$ controls the decay of the semantic score when the PA increases. In our experiments, we used $\lambda = 5$.
 
\paragraph{Video compositing.} In this step, our methodology removes frames from the original video using the semantic curve. Firstly, for each frame of the video, we assigned a semantic score concerning the sound segments. Then, according to the scores of each frame, the video is segmented into semantic and non-semantic parts, where a semantic segment has low PA values. After the segmentation, we compute for each segment a different speed-up proportional to the semantic level of the segment. Thus, segments with higher semantic levels receive a lower speed-up whilst respecting the overall required speed-up. 

To select which frames to remove, we use the Multi-Importance Fast-Forward (MIFF) method, the state-of-the-art semantic-aware hyperlapse method~\cite{Silva2018}. The MIFF method is based on a directed graph, where each node represent the frames of the video and edges representing the transition between pair frames. A weight is attributed to each edge of the graph according to metrics as shakiness, visual appearance, the speed of motion and semantics (in our case, the PA of the segment that the frame is part of). Then, the method generates the shortest path of each graph and concatenates these paths. In the last step, the MIFF method stabilizes the concatenate path generating a more visually pleasant fast-forward video.
\section{Experiments}

\paragraph{Experimental setup.} To evaluate our method we used the Dataset of Multimodal Semantic Egocentric Videos (DoMSEV) proposed by Silva~\etal~\cite{Silva2018CVPR}. This dataset is composed of $80$ hours egocentric videos. It provides videos, sound, IMU measurements, GPS, and depth. Each video has annotations specifying the ambient where the video was recorded (indoor, outdoor), the activity performed, and recorder attention/interaction during the video.

We manually selected $11$ segments from videos in the DoMSEV dataset with a rich diversity of sound and visual events (e.g., crowded places and quiet streets and parks). For each segment, we split its soundtrack into segments of $3$ seconds. Regarding the semantic hyperlapse creation, all parameters were set with the same values used by the authors in their experiments~\cite{Silva2018}.

\paragraph{Evaluation Metrics and baseline.} We used two metrics to evaluate the results of the experiments: the speed-up rate and the instability index~\cite{Silva2016}. The Speed-up rate describes the ratio of frames between the input and the output video. The instability index measures the amount of instability generated by the method in the resultant video.

We pit our method against the MIFF method~\etal~\cite{Silva2018}, where each frame receives a semantic score according to the number of detected faces and pedestrians on the image.

\paragraph{Results.}

\label{sec:experiments}
\begin{figure}[t]
\centering
\includegraphics[width=0.45\textwidth]{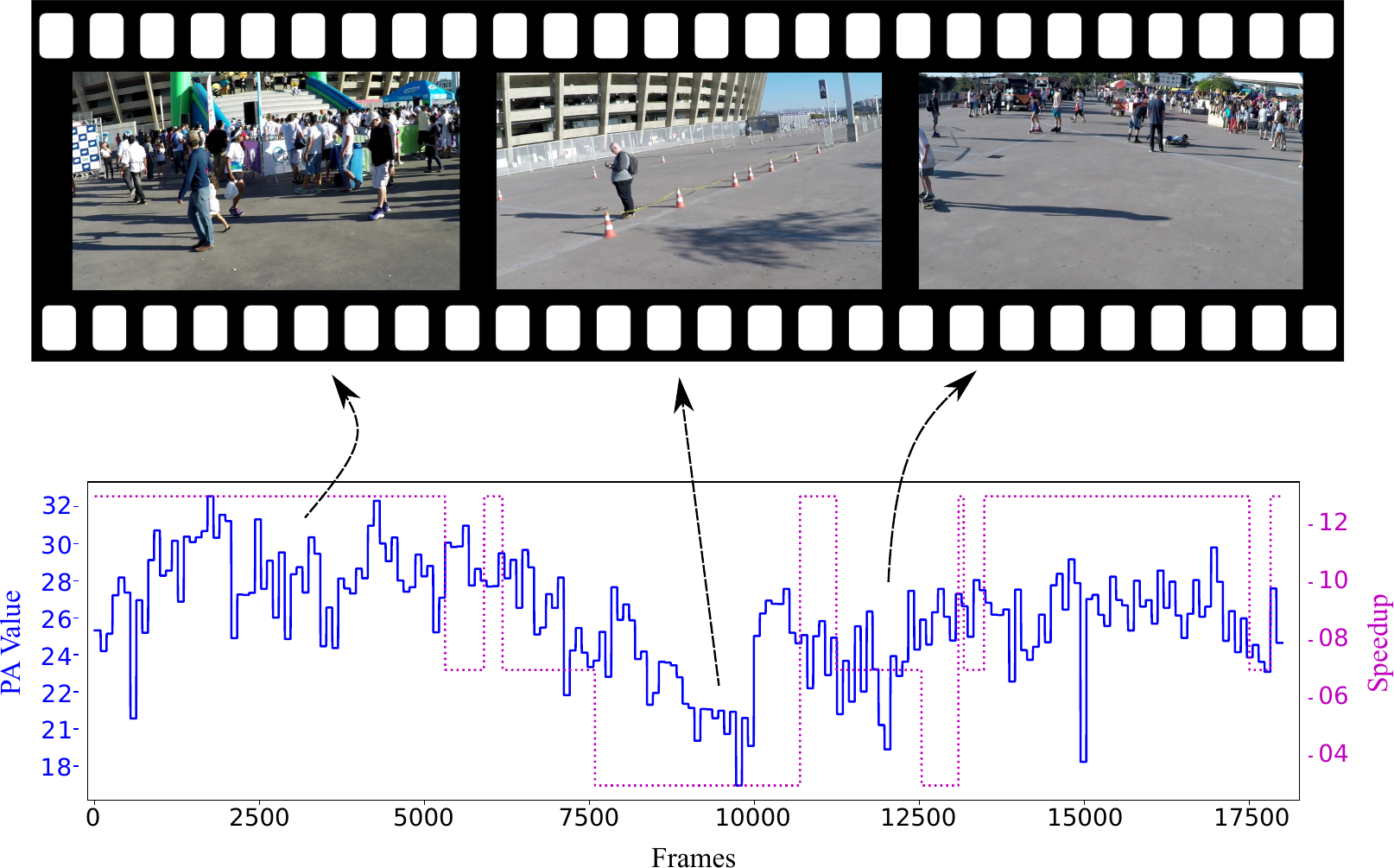}
\caption{A qualitative analysis of the speed-up rates and PA values. On the left side, a frame from a segment with high PA value. In the middle a frame from a segment with a low PA and on the right a frame from a segment with medium PA value.}
\label{fig:visualresult}

\end{figure}

Figure~\ref{fig:visualresult} shows the visual results of our method. The blue curve represents the semantic score, and the pink curve shows the speed-up value. We highlighted three frames on the figure representing different segments of the video with different PA values. The first frame shows a segment with a high PA that received a high speed-up. We can see several sources of acoustic annoyance in the image as a soundbox near to the recorder and a crowd. The second frame is related to a low value of psychoacoustic annoyance and happens when the recorder goes away from the crowd and the music playing in the background. This segment received the lowest speed-up in the video. In the third part, the record decides to go back to the crowd. Thus, we have a medium value of PA, since loudness and sharpness metrics increase when close to the crowd and music.

Table~\ref{tab:pleasantness_results} presents the quantitative results. It contains the Instability Index~\cite{Silva2016}, speed-up, and the average PA achieved by each method in the $11$ segments of video used in the experiments. Regarding speed-up, our method achieved the desired factor ($10\times$) in most of the videos. It presented a better result when compared to MIFF method. MIFF performs slightly better than our method regarding instability. One reason may be the large number of speed-up thresholds representing different levels of semantic importance created because of granularity of the semantic information extracted by our method. For this reason, the value of the speed-up increases, generating a higher discrepancy between frames of the fast-forward version of the videos. When considering the average PA of the final video, our method produced fast-forwarded videos with lower mean PA than the MIFF method.

\begin{table}[t!]
	\caption{Speed-up and instability results for the $11$ videos segments used. We used a $10\times$ speed-up factor.}
	\label{tab:pleasantness_results}
	\setlength{\tabcolsep}{3.5pt}
	\small{
		\begin{tabular}{lrrrrrrrrrrrllcccc} \toprule
			& \multicolumn{2}{c}{\textbf{Speed-up}$^1$}   & & \multicolumn{2}{c}{\textbf{Instability}$^2$} & & \multicolumn{2}{c}{\textbf{Mean PA}$^2$} \\  
			\thead{\textbf{Videos}}             & \thead{Ours}          & \thead{MIFF}     & & \thead{Ours}          & \thead{MIFF} & & \thead{Ours}          & \thead{MIFF} \\ \cmidrule(l){2-3} \cmidrule(l){5-6} \cmidrule(l){8-9} \\
			Academic\_Life\_03 & 10.0 & 10.0 & &  39.4 &  34.6 & &  20.5 &  21.5 \\
			Attraction\_01     & 10.0 & 10.0 & &  37.5 &  37.5 & &  20.9 &  21.8 \\
			Daily\_Life\_01    & 10.0 & 15.8 & &  49.0 &  50.0 & &  21.4 &  22.0 \\
			Party\_01    	   & 10.0 & 10.0 & &  29.6 &  29.0 & &  28.1 &  28.3 \\
			Recreation\_01     & 10.0 & 10.0 & &  41.0 &  37.8 & &  13.8 &  15.1 \\
			Recreation\_02     & 10.0 & 10.0 & &  45.0 &  40.1 & &  23.5 &  24.7 \\
			Recreation\_04     & 10.0 & 10.0 & &  38.9 &  33.3 & &  25.8 &  27.4 \\
			Recreation\_06     & 10.0 & 12.3 & &  38.6 &  42.4 & &  22.0 &  22.9 \\
			Recreation\_08     & 10.0 & 10.2 & &  27.9 &  27.8 & &  24.2 &  24.8 \\
			Shopping\_01       & 10.0 & 10.2 & &  41.5 &  37.1 & &  22.0 &  23.3 \\
			Tourism\_02        & 10.0 & 12.4 & &  39.8 &  40.0 & &  22.3 &  23.2 \\
			\cmidrule(l){2-3} \cmidrule(l){5-6} \cmidrule{8-9}
			\textit{Mean} 	   & \textit{\textbf{10.0}} & \textit{11.0} & &  \textit{38.9} &  \textit{\textbf{37.2}} & &  \textit{\textbf{22.2}} &  \textit{23.2} \\ 
			& \multicolumn{4}{l}{\scriptsize{$^1$\textit{Better closer to 10.}}} &  \multicolumn{4}{l}{\scriptsize{$^2$\textit{Lower is better.}}} \\ \bottomrule
		\end{tabular}
	}
\end{table}

\section{Conclusion}

In this paper, we proposed a new approach to extract semantic from the soundtrack of egocentric videos. This semantic is combined with visual features for fast-forwarding the input video creating a smaller and stabilized new version. We estimate the annoying sound moments from the video applying psychoacoustic metrics. In our results, we show that even though we lost some stability when compared with state-of-the-art methods, our method can fast-forward a video using the sound information and prevent annoying acoustic segments in the final video.
\paragraph{Acknowledgments.}

The authors would like to thank CAPES, CNPq, FAPEMIG, Petrobras, and NSF VeHICaL project ($\# 1545126$) for funding this work.

{\small
\bibliographystyle{ieee}
\bibliography{CVPRW_2018}
}

\end{document}